\documentclass[ sn-mathphys,Numbered]{sn-jnl}% Math and Physical Sciences Reference Style
\usepackage{tabulary}
\usepackage{graphicx}%
\usepackage{multirow}%

\usepackage{amsmath,amssymb,amsfonts}%
\usepackage{amsthm}%
\usepackage{mathrsfs}%
\usepackage[title]{appendix}%
\usepackage{xcolor}%
\usepackage{textcomp}%
\usepackage{manyfoot}%
\usepackage{booktabs}%
\usepackage{algorithm}%
\usepackage{algorithmicx}%
\usepackage{algpseudocode}%
\usepackage{listings}%
\usepackage{array}
\usepackage{textcomp}
\usepackage{multirow}
\usepackage{graphicx}
\usepackage{subcaption}
\usepackage{colortbl}
\usepackage{tabularx, float, makecell}
\usepackage{xcolor}
\usepackage{soul}

\usepackage{adjustbox}
\def\BibTeX{{\rm B\kern-.05em{\sc i\kern-.025em b}\kern-.08em
    T\kern-.1667em\lower.7ex\hbox{E}\kern-.125emX}}
\usepackage{hyperref}
\hypersetup{
    colorlinks=true,
    linkcolor=magenta,
    filecolor=forestgreen,      
    urlcolor=cyan,
    pdftitle={Overleaf Example},
    pdfpagemode=FullScreen,
    }

\usepackage{marginnote}
\usepackage{soul}

\theoremstyle{thmstyleone}%
\theoremstyle{thmstyletwo}%

\theoremstyle{thmstylethree}%

\begin{document}
\title{\LARGE \bf
Levels of explanation- implementation and evaluation of \textbf{\textit{what}} and \textit{\textbf{when}} for different time-sensitive tasks  
}

%%=============================================================%%
%% Prefix	-> \pfx{Dr}
%% GivenName	-> \fnm{Joergen W.}
%% Particle	-> \spfx{van der} -> surname prefix
%% FamilyName	-> \sur{Ploeg}
%% Suffix	-> \sfx{IV}
%% NatureName	-> \tanm{Poet Laureate} -> Title after name
%% Degrees	-> \dgr{MSc, PhD}
%% \author*[1,2]{\pfx{Dr} \fnm{Joergen W.} \spfx{van der} \sur{Ploeg} \sfx{IV} \tanm{Poet Laureate} 
%%                 \dgr{MSc, PhD}}\email{iauthor@gmail.com}
%%=============================================================%%

\author*[1]{\fnm{Shikhar} \sur{Kumar}}\email{shikhar@post.bgu.ac.il}
\equalcont{These authors contributed equally to this work.}

\author[1]{\fnm{Omer} \sur{Keidar}}\email{omerkei@post.bgu.ac.il}
\equalcont{These authors contributed equally to this work.}

\author[1]{\fnm{Yael} \sur{Edan}}\email{yael@bgu.ac.il}

\affil*[1]{\orgdiv{Department of Industrial Engineering and Management}, \orgname{Ben-Guion University of the Negev}, \orgaddress{ \city{Beer Sheva}, \country{Israel}}}

%%==================================%%
%% sample for unstructured abstract %%
%%==================================%%

\abstract{As robots become more and more capable and autonomous, their usage in daily tasks by nonprofessional users and bystanders will increase.
For smooth and efficient human-robot interaction (HRI), robots should be designed such that their planning, decisions, and actions are understood by the humans with whom they interact.  
This can be achieved by having the robot explain its actions.
In this work, we focused on constructing and evaluating levels of explanation(LOE) that address two basic aspect of HRI:
1. \textit{What} information should be communicated to the user by the robot? 
2. \textit{When} should the robot communicate this information? 
For constructing the LOE, we defined two terms, verbosity and explanation patterns, each with two levels (verbosity - high and low, explanation patterns - dynamic and static).
Based on these parameters, three different LOE (high, medium, and low) were constructed and evaluated in a user study with a telepresence robot. 
The user study was conducted for a simulated telerobotic healthcare task with two different conditions related to time sensitivity, as evaluated by two different user groups - 
one that performed the task within a time limit and the other with no time limit. 
We found that the high LOE was preferred in terms of adequacy of explanation, number of collisions, number of incorrect movements, and number of clarifications 
when users performed the experiment in the \textit{`without time limit’} condition.
We also found that both high and medium LOE did not have significant differences in completion time, the fluency of HRI, and trust in the robot. 
When users performed the experiment in the \textit{`with time limit’} condition, high and medium LOE had better task performances and were preferred to the low LOE in terms of completion time, fluency, adequacy of explanation, trust, number of collisions, number of incorrect movements and number of clarifications. 
Future directions for advancing LOE are discussed.}


\maketitle
\section{Introduction}
As robots become more and more capable and autonomous, 
their implementation in daily tasks by nonprofessional users and bystanders will increase.
%Indeed, the increased deployment of several types of robotic platforms during the COVID-19 pandemic illustrates the paradigm shift in this field~\cite{tamantini2021robotic}.
The widespread penetration of robotics platforms into non-industrial settings  
and the increasing trend of deployment of collaborative robots~\cite{sadangharn2022multidimensional,portugal2019study}, 
indicate that increased interaction with humans will be required~\cite{lambert2020systematic}. 
Importantly, for smooth and efficient interaction, the robots should be understandable to the interacting humans~\cite{chiou2021towards} or, in other words, the planning, decisions, and actions of robots should be understood by the interacting humans~\cite{fong2002survey}.
An understandable robot will also improve the user’s perception of the robot~\cite{bensch2017interaction}, its usability~\cite{ baud2014human}, and acceptance~\cite{ye1995impact}.   
A robot’s inability to explain its ‘thinking’ or actions could even lead to anxiety on the part of the interacting human~\cite{nomura2011relationships}. 

A theoretical model of understandable robots~\cite{hellstrom2018understandable} claims that a communicative action (verbal or non-verbal explanations by the robot to the user) should be generated when there is a discrepancy between the robot’s state of mind and the human’s perception of the robot’s theory of mind. 
They proposed that such explanations should be guided by the following questions: 
\begin{enumerate}
    \item \textit{\textbf{What}} information (if any) should be communicated to the human? 
    \item \textit{\textbf{How}} should the robot represent and infer the human’s mind?
    \item \textit{\textbf{How}} should communicative actions be generated 
to communicate the required information? 
    \item \textit{\textbf{To whom}} should the robot direct the communicative actions? 
 \item \textit{\textbf{Which mechanism}}
should enable the model to focus on two sub-questions related to \textit{\textbf{if}} and \textit{\textbf{when}}.
\end{enumerate}

The current paper focuses on establishing the basis for the verbalization of explanations by composing levels of explanation (LOE).
The LOE which were designed on the basis of the above-mentioned~\cite{hellstrom2018understandable} study take into account the interaction design and applicability as detailed below. 
To avoid an evaluation of a complex design
with all four questions (\textbf{what},\textbf{when}, \textbf{why}, \textbf{how}),
in this work we focused only on two questions (\textbf{what} and \textbf{when}). 
We constructed three levels of explanation (LOE) for  a robot and evaluated these LOE in a telerobotics user study. 

%We  note that a telerobotics, being a robot that is controlled remotely by a human operator~\cite{evans2018telemedicine,zhang2022telepresence}, can provide the user with different opportunities to interact directly with the environment, even if there are barriers that would prevent her/him from doing so physically~\cite{siciliano2008springer}.

An important aspect of HRI that must also be taken into consideration in the design of explanations is the task criticality~\cite{yanco2002taxonomy}, which is defined as \textit{``the importance of getting the task done correctly in terms of negative effects should a problem occur."}
In some cases (like search-and-rescue), critical tasks should be executed within a specific time limit~\cite{drury2006applying}, 
i.e., some tasks are also time sensitive.
A key component of an explanation for such a task would be the time available to the user to listen to the robot's explanation~\cite{papagni2021understandable}. 
The amount of time that the user can spend on listening depends on the environment and the task.
For example, in a library environment, users will probably have time to hear detailed explanations from a robot regarding suitable reading material~\cite{sreejith2015conceptual}. 
In contrast, in search-and-rescue operations, 
 a brief and precise explanation would be required~\cite{7451740} to accomplish a task since the user must perform the task in a limited time. 

In this paper,  we focused on a teleoperation navigation task. 
Since real-world operation must ensure safety in a dynamic environment, 
the control of the robot may be handed over to the human when the robot is required to maneuver around obstacles.
We focus on explaining the planning or ‘thinking’ of the robot to the user  
(i.e., explaining \textbf{what} the robot is doing) and also address the particular time at which explanations must be communicated 
by the robot (\textbf{when})  to ensure an efficient interaction.

%To evaluate the influence of time sensitivity on the levels of explanation, 
%we compared two conditions, 
%one \textit{’with time limit’} and the other \textit{’without time limit’}. 

\section{Related Works}
A theoretical model for generating explanations mapped the cognitive AI process to the human social process; this model proposed five levels of explanation~\cite{dazeley2021levels}, i.e, (1) interpreting a decision based on the available information; 
(2) explaining the internal functions of agents (for example, robots) interacting with other actors (for example, interacting humans) in the environment;
(3) explaining the internal functions based on the beliefs or mental states of other agents (for example, interacting humans);
(4) interpreting the systems based on the cultural expectations of the users;
(5) explaining all the above factors that were responsible for generating the explanations.
The study placed greater emphasis on aligning the explanations based on human cognitive abilities; it did not address the verbalization of the explanations.
The need for verbalization of the explanation is influenced by Grice's maxim of quantity~\cite{grice1975logic}.
It is described as the optimal information that must be communicated for better understanding.

Previous research has referred to the different modalities in which (\textbf{How})
explanations need to communicated (e.g.,~\cite{han2021need,adamik2022explainability}). 
In~\cite{han2021need}, three different modalities were used for explanations, i.e., textual, graphical, or both.
It was revealed that the graphical method of explanation was preferred.
Another study demonstrated that there is a need for explanations from the robot irrespective of expected or unexpected behavior by the robot \cite{adamik2022explainability}.
That study further elaborated on the importance of verbal explanations and demonstrated that verbal explanations accompanied by non-verbal explanations improved user perception. 
The importance of verbal communication in combination with non-verbal communication was supported by another study~\cite{tiferes2019gestures} in which it enhanced the team communication between human-human interaction while working with robot-assisted surgery.
%Therefore, this study focuses on the development of LOEs to improve the verbalization of the explanations.
Yet another study, proactive explanations~\cite{iio2020human}  provided by a robot in a field setting (museum) were perceived positively.
These studies revealed that explanations provided by the robot are necessary for interaction between humans and robots.

In a study that developed the verbalization of multi-robot collaboration~\cite{raman2013sorry,singh2021verbal,korpan2017natural}
based on Grice's maxim of quantity, the verbal utterances increased the understandability of the robot to bystanders.
However, that study did not present different levels of verbalization of robot utterances.
Another study~\cite{das2021explainable} dealing with robotic failures proposed four different types of explanation, i.e., ``action-based" (explanations provided only for the internal state of the robots), ``context-based" (explanations also include environmental factors in addition to ``action-based"), ``action-history based" (explanations involve historical explanations of a particular action in addition to ``action-based"), and ``context-historical based" (involve additional historical aspects to the ``context-based").
Results revealed that  ``context-historical based" explanations aided failure recovery.
The content of the explanation was limited to \textbf{What} information needed to be communicated.

In~\cite{8673198}, inverse reinforcement learning was used to predict the user’s action and to prevent the failure of a task by generating explanations in two modes.
The first mode involved only informing the user about a particular action that would lead to failure. 
The second mode involved explaining \textbf{why} (the reason(s) for the failure) a particular action would lead to failure of a task. 
It was found that explaining the reason(s) for failures improved user performance.
In~\cite{zhu2020effects}, proactive explanations (explaining the robotic actions, \textbf{What}, as well as their justification, \textbf{why}) were perceived positively as compared to a proactive announcement alone (explaining only the robotic action).

In~\cite{chiou2021towards}, four different explanations and strategies were evaluated for search-and-rescue operations: 
always explain (pushed information), explain on-demand of the user, information (which was always available) pulled by the user, and never explain.
The best strategy was to moderate the information pushed by the robot.
A similar finding was supported by another study~\cite{keidar2022push} in which pushing information resulted in improved task performance~\cite{Inertactionmodes2022}.
Accordingly in our design, the information is pushed to the user. 
In another study, explanations were generated utilizing a model reconciliation algorithm~\cite{SREEDHARAN2021103558}.
However, that study did not focus on the verbalization of explanations. 
We developed a model for verbalization of the explanations, as explained in the next section.  
The content of the explanations includes both \textbf{What} and \textbf{When}.

\section{Levels of Explanation}
The proposed LOE were inspired by the design of levels of automation (LOA)~\cite{vagia2016literature},
with the basic scheme for  LOE based on classification themes that have been proposed for LOA~\cite{endsley1997use,endsley1999level,mcdaniel1988rules,parasuraman2000model,rouse1991design,sheridan1992supervisory,sheridan1978human}.
Such classification schemes~\cite{jipp2014levels} allowed the implementation of automation systems with different  LOAs, enabling their empirical evaluation~\cite{endsley1995out,kaber2000design,manzey2008performance,moray2000adaptive}. 
Similar empirical evaluations are critical for advancing the field of HRI~\cite{hoffman2020primer}.

For practical purposes, it is important to present to the user only a small number of levels 
(\cite{olatunji2020user}).
 In this research, we propose a basic approach for LOE involving only two aspects of understanding, 
 \textbf{What} and \textbf{When}~\cite{hellstrom2018understandable}.
 We focus on two questions:
 \begin{itemize}
     \item \textbf{What} information should be communicated to the user by the robot?
     \item \textbf{When} should the robot communicate the information, i.e., should the robot explain its plan before putting the plan into action or during the implementation of the plan? 
 \end{itemize}
 
 The rationale underlying the construction of these LOE can be explained through the use of two examples, as follows. 
In a search-and-rescue task information should be communicated in a concise manner and at the right time.
 Otherwise, it would be challenging for the user to decipher a multitude of information and s/he could miss important information in the task if it is not provided at an appropriate time. 
In a demining task, it is important to provide the user with detailed information to ensure correct and accurate performance of the task; 
the robot should explain its plan before starting its execution and along the implementation of the plan the robot should supervise and guide the user throughout all the steps of the task. 

Two terms were defined to address the \textbf{What} and \textbf{When} questions,  verbosity and explanation patterns. 
 Verbosity is defined as the amount of information that must be communicated to the user for a better understanding of the robot (i.e., \textbf{What}). 
 We divided verbosity into two levels, high and low. 
 At a high level of verbosity, the robot explains all the details of the action that it has planned. 
 For example, in a telerobotics scenario, a robot facing an obstacle would explain to the user that it needs to turn 90° to the right, then move forward, and so on.
 At the low level of verbosity, the robot would explain only the broad outline of the plan without any details; for example, 
 it would explain only that it needs to turn right without detailing how this maneuver would be accomplished. 

The explanation pattern was 
defined by whether the communicative actions are delivered as a one-time ’package’ or 
whether the delivery is ongoing with the task (i.e., \textbf{When}). 
Accordingly, we defined two explanation patterns, 
static (one time) and dynamic (ongoing). 
In the static explanation pattern, the robot explains the plan to the user only once before the execution of the plan, with the explanation depending on the level of verbosity.
In the dynamic explanation pattern, the robot explains each action in parallel to its execution, again according to the level of verbosity. 

\begin{figure}[h!]
\centering
{%
\resizebox*{8cm}{!}{\includegraphics{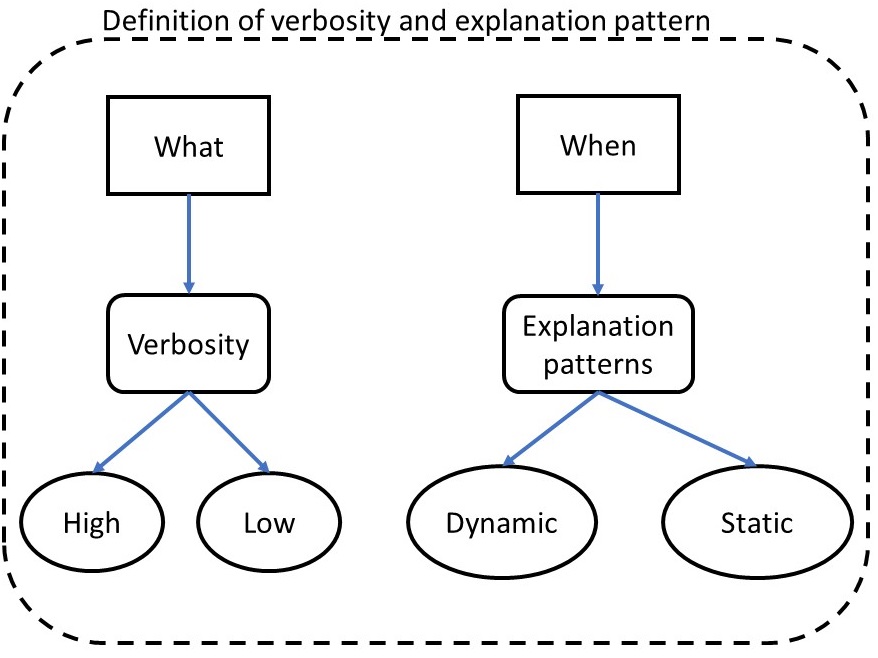}}}\hspace{2pt}
{%
\resizebox*{8cm}{!}{\includegraphics[width=\linewidth]{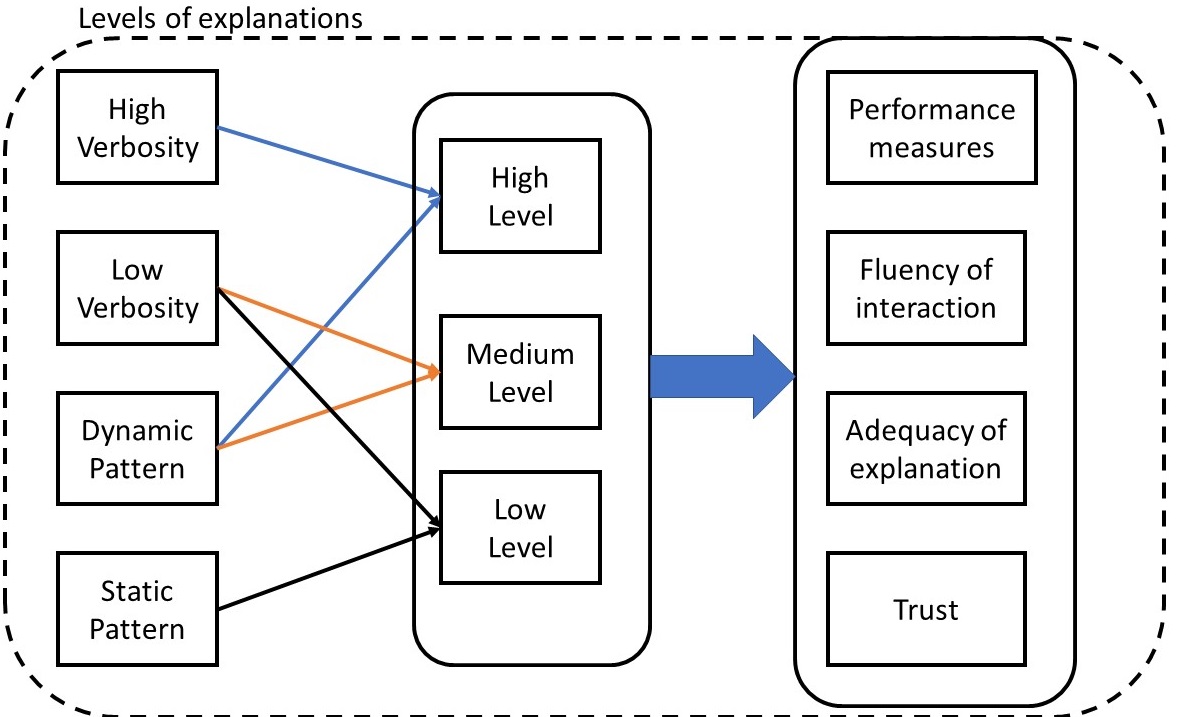}}}
\caption{Definition of levels of explanation for \textbf{What} and \textbf{When}.}
\label{fig:model}
\end{figure}

Based on the above definitions, three LOE (high, medium, and low) were configured, as shown in Figure \ref{fig:model}: the high LOE has high verbosity and a dynamic pattern; the medium LOE has low verbosity and a dynamic pattern; and the low LOE has low verbosity and a static pattern. 

The decision to use only three LOE was based on previous recommendations, which noted the inability of users to differentiate between more than three levels~\cite{olatunji2020user}. 
This configuration also ensured a simple and feasible experimental setup. 
We did not employ the combination of a high level of verbosity and a static pattern of explanation because a detailed explanation in a static pattern could have increased the user's cognitive load. 

\subsection{Research hypotheses}
The LOE model (Fig. \ref{fig:model}) is based on the following hypotheses. 

We assume that in situations in which time is restricted, the user would require brief explanations~\cite{7451740}. 
We posit that the limited information provided by the robot will affect the completion time~\cite{gunning2019xai} and other performance metrics, the fluency of the HRI~\cite{8678448}, the adequacy of the explanation~\cite{hoffman2018metrics}, and trust~\cite{gunning2019xai}. 
Hence, we propose Hypothesis 1 under time constraint conditions as follows:
\begin{itemize}
    \item[\textit{H1.1:}] \textit{Participants will perceive medium LOE positively as compared to the other two LOE}.
    \item[\textit{H1.2:}] \textit{Participants will perform better in medium LOE as compared to the other two LOE}.
\end{itemize}

We also assume that when there is no time limit the user requires more information when interacting with the robot~\cite{sreejith2015conceptual,mcneese2018teaming}.
We posit that more information provided by the robot to the human would affect the completion time~\cite{gunning2019xai} and other performance metrics, the fluency of interaction~\cite{8678448}, the adequacy of explanation~\cite{hoffman2018metrics}, and trust~\cite{gunning2019xai}.
Hence, we propose Hypothesis 2 when operating under no time constraint conditions, as follows: 
%a high LOE will yield the following as compared to the other two LOEs:
\begin{itemize}
    \item[\textit{2.1:}]\textit{Participants will perceive high LOE positively as compared to the other two LOE.}
    \item [\textit{2.2:}]\textit{Participants will perform better in high LOE as compared to the other two LOE}
\end{itemize}
%\textit{H2.1: A shorter completion time.}
%\textit{H2.2: More fluent interaction.}
%\textit{H2.3: Adequate explanations.}
%\textit{H2.4: Higher trust.}

\section{Methods}
\subsection{Overview}

The three LOE were implemented and evaluated in a teleoperation task.
The task was conducted under two different time-sensitive conditions (termed task condition throughout the paper), 
\textit{`with time limit'} and \textit{`without time limit.'} 
Participants were voluntarily recruited and
divided into two equal groups randomly according to the task conditions. 
In the first task condition, users performed the task without any time constraints. 
In the second task condition, users were 
required to accomplish the human-robot collaborative task within a specified time limit.

The experiment was designed to be a within-the-group design with all three LOE as independent variables (each user experienced all three LOE).
Dependent variables included subjective measures 
(human-oriented fluency, adequacy of explanations, and trust ), as shown in Table~\ref{tab:FinalQ} and objective measures (completion time, number of collisions, incorrect movements, number of clarifications), as shown in Table~\ref{tab:objective}.

\subsection{The experimental system}
The experimental system consisted of a mobile robot platform, remote user interfaces, and a server-client communication architecture that was connected to the robot operating system (ROS) platform via a Rosbridge WebSocket. The user interfaces were run on the teleoperator’s computer and were programmed in HTML, JS, CSS, and PHP. 

The mobile robot platform was a 1.64-m high Keylo telepresence robot with a low center of gravity and a circular footprint of 52 cm in diameter. 
The Keylo is equipped with a 24" FOV touchscreen. 
It runs on Ubuntu 18.04 LTS and ROS Melodic with a standard ROS API for all sensors and features.
The navigation sensors include a LiDAR sensor (Hokuyo URG-04LX-UG01, range 5.6 m, FOV 240°), two sets of four front and rear ultrasonic range sensors (5 m range), and two sets of two  IR edge detectors hard-wired to the motor's controller. Additionally, the robot is equipped with three 3D RGB-D Intel RealSense™ R200 cameras, two front, and one rear.

\subsection{Design of the interfaces}
Two interfaces were built, 
one for each task condition 
(\textit{`with time limit'} and \textit{`without time limit'}). 
The interfaces were designed to be user-centered, 
in line with previous research recommendations~\cite{olatunji2020user}. 
Based on previous research findings, 
the interface design rested on a proactive interaction mode in which all the information was `pushed’~\cite{Inertactionmodes2022}, 
with both visual and auditory feedback~\cite{markfeld_2020,gutman2023evaluating}. 

\begin{figure}[h]
\centering

\subfloat[\textit{`without time limit'} condition]{\includegraphics[width=\linewidth]{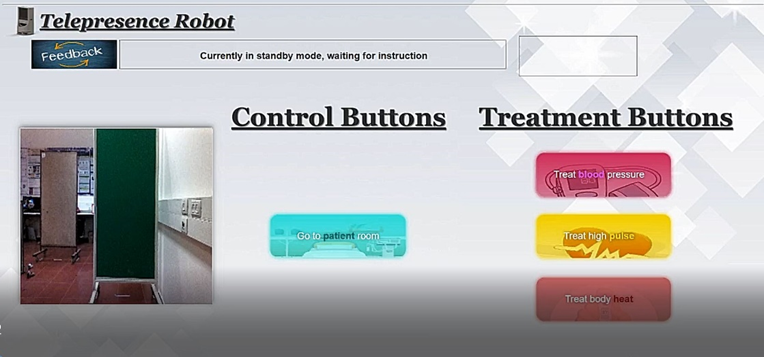}\label{WithoutLimitInterface}}
\hspace{0.1cm}
\subfloat[\textit{`with time limit'} condition]{\includegraphics[width=\linewidth]{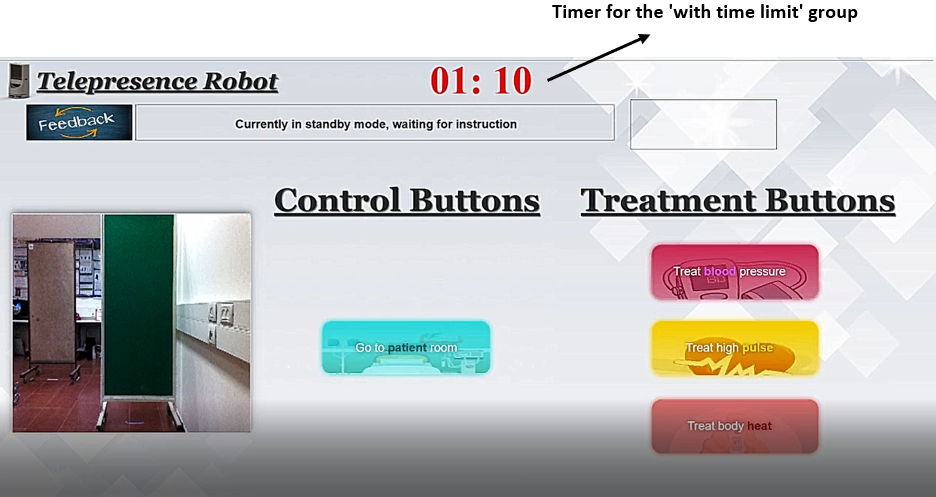}\label{WithLimitInterface}}
\caption{Design of the interface that was used by the participants to navigate the robot.}
\end{figure}

The interfaces included a display of the front camera view, feedback, and explanations from the robot (e.g., explanations for bypassing obstacles, reaching important points, and warnings about obstacles), and four clickable `buttons' - one for sending the robot automatically to the locations in the task, i.e., a ``Go to patient room" button, and three buttons for treating the patient in accordance with the task.
Manual control of the robot was executed via the arrow keys on the keyboard (Fig.~\ref{WithoutLimitInterface}). 

In the interface of the \textit{`with time limit'} task condition, the time limit was indicated by a red digital timer located at the top of the interface (Fig. \ref{WithLimitInterface}).
A specific time limit of one minute and ten seconds was empirically determined in several pilot runs.
The timer was started each time an obstacle was reached and then reset after the obstacle had been bypassed.

%%%  style for figure starts here
% Figure

\subsection{Participants}
We recruited 30~\footnote{The required sample size of 27 was calculated with the
Gpower software~\cite{erdfelder1996gpower} for an 
effect size of $0.25$, $\alpha=0.05$, and power of $0.95$ with a test family of F-test 
with three groups (number of independent variables) corresponding to the number of LOEs and seven metrics.} 
students for each task condition (a total of 60 participants ) via a call for volunteers to take part in the experiment in the Department of Industrial Engineering and Management, Ben-Gurion University of the Negev. 
The volunteers were offered a bonus point for the final grade in a compulsory undergraduate course, commensurate with the time spent on the experiment. 
The experiment was approved by the Department's ethical committee.

The participants were assigned at random into one of the two groups while ensuring a gender balance: 
\textit{`with time limit'} (15 males and 15 females; mean age = 26.1, SD = 1.32) and \textit{`without time limit'} (15 males and 15 females; mean age = 26, SD = 1.31).

All participants had experience in programming robots but no previous experience with operating teleoperated robots.
We took students to ensure a homogeneous population with similar backgrounds and to avoid the novelty effect~\cite{reimann2023social}. 

%\subsection{Experimental setting}
%The experimental setting was arranged to resemble a clinic that housed three patients and five obstacles (Fig. \ref{cross-section}). Each patient was simulated by a monitor on which the patient’s vital signs (in this experiment blood pressure, body temperature, and pulse) were visible. The obstacles were simulated by poster stands. 

%The three LOEs (high, medium, and low) were examined in the two different conditions, i.e., by comparing the performances of the two groups, \textit{`with time limit'} and \textit{`without time limit'}. The influence of the LOE on different aspects of user performance and user perception was evaluated.

%%%  style for figure starts here
% Figure

%%%  style for figure ends here

\subsection{The task}

\begin{figure}[h!]
\centering
\includegraphics[scale=0.6]{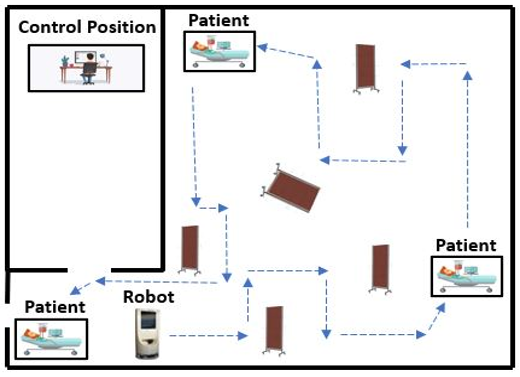}
\caption{Cross-section of the experimental environment}\label{cross-section}
\end{figure}

During the task, the remote operator (user), seated at the controls as shown in Fig. \ref{cross-section}, was required to navigate the robot in an environment simulating a clinic such that s/he would virtually visit each of three patients to monitor their vital signs and administer treatment if necessary, as described below. 
The user was given the range of normal values for the vital signs at the beginning of the experiment. 
If the user found that any of the vital signs did not fall in the normal range, then s/he was required to `treat' the patient by clicking on the options provided on the interface.

The task included three obstacles and three patients such that each user experienced all three different LOE in the same task (each obstacle and patient were defined as one LOE). 
In each experiment, the robot advanced autonomously along a path that was pre-defined at the beginning of the task until it reached an obstacle. 
At that stage, it stopped and informed the user about the presence of the obstacle and its inability to bypass the obstacle. 
The robot then gave instructions to the user as to how to teleoperate it around the obstacle, according to the particular LOE for that obstacle and that patient, as shown in Table~\ref{tab: ExplanationsOfLOEAPPENDIX}. 
Once the user had succeeded in teleoperating the robot around the obstacle (according to the relevant LOE), 
the robot informed him/her of the success of the bypass maneuver and then automatically continued to the patient.
When the robot reached the patient, the user was able to see the patient’s metrics through one of the robot's cameras. 
At that stage, the user either received or did not receive an explanation from the robot regarding the need to check the patient's metrics, depending on the particular LOE (Table~\ref{tab: ExplanationsOfLOEAPPENDIX}), and accordingly, was required to administer the appropriate treatment. 
Only after the user had administered the correct treatment to the patient (different treatments were assigned for each LOE) did the robot automatically move on toward the next patient (next LOE).

The difficulty of the obstacles was identical for all three LOE (same distance for the robot to navigate, same distance between the patient and the next obstacle, and the same number of movements to be performed to bypass the obstacle).
The participants were free to request help from the experimenter at any stage of the experiment.

%%%%%%%%%%%%%% START OF TABLE %%%%%%%%%%%%%%

\begin{table*}[h!]
\centering

\caption{Explanations provided by the robot to the user in the form of instructions with the following dialogs being generated by the robot in each LOE.}
\label{tab: ExplanationsOfLOEAPPENDIX}
\resizebox{\linewidth}{!}{%
\begin{tabular}{|l|l|}
\hline
\textbf{Level of Explanation} & \textbf{Explanations from the robot}   \\ \hline
High                          & \begin{tabular}[c]{@{}l@{}}A very detailed explanation regarding the passage of the obstacle - each movement that the \\ participant had to perform to bypass the obstacle was detailed separately (the participant received \\ the next explanation only after completion of the action in the previous explanation):\\    \\ 1. ``Turn left 90 degrees"\\ 2. ``Move forward"\\ 3. ``Stop and turn right 90 degrees"\\ 4. ``Move forward"\\ 5. ``Stop and turn right 90 degrees"\\ 6. ``Move forward"\\ 7. ``Stop and turn left 90 degrees"\\ 8. ``Move forward"\\    \\ After bypassing the obstacle: ``You have bypassed the obstacle, continue to the patient"\\ When reaching the patient: ``I'm in the patient's room, check patient's metrics"\\ When the participant has finished giving appropriate treatment to the patient:\\ ``On the way to the next patient's room"\end{tabular} \\ \hline
Medium                        & \begin{tabular}[c]{@{}l@{}}A partially detailed explanation regarding the passage around the obstacle - an explanation that \\ combined the movements that are required (the participant received \\ the next explanation only after completion of the action in the previous explanation):\\    \\ 1. ``Turn left and move forward"\\ 2. ``Stop, turn right and move forward"\\ 3. ``Stop, turn right and move forward"\\ 4. ``Stop, turn left and move forward"\\    \\ After bypassing the obstacle: ``You have bypassed the obstacle, continue to the patient"\\ When reaching the patient: The robot gave no explanation.\\ When the participant has finished giving appropriate treatment to the patient:\\ ``On the way to the next patient's room"\end{tabular}                                                                                                                \\ \hline
Low                           & \begin{tabular}[c]{@{}l@{}}A single explanation of all the movements that need to be done to bypass the obstacle:\\    \\ 1. ``Turn left, then turn right and then right again and finally turn left"\\    \\ After bypassing the obstacle: ``You have bypassed the obstacle"\\ When reaching the patient: The robot gave no explanation.\\ When the participant has finished giving appropriate treatment to the patient:\\ ``On the way to the next patient's room"\end{tabular}   \\ \hline
\end{tabular}%
}
\end{table*}

%%%%%%%%%%%%%% END OF TABLE %%%%%%%%%%%%%%
%\subsection{Experimental design}
%The experiment was designed as a within-between experiment with the LOE as the within independent variable and the time limit as the between independent variable. 
%Each participant was randomly assigned to one of the two groups (taking into consideration the need to maintain a 50:50 M:F ratio in each group), 
%and the order in which the participant was required to execute the different LOEs was randomly determined.
%Accordingly, there were six different sequences for the task, and 5 participants in each group performed the task in each sequence.

%%%% END OF TABLE

\subsection{Procedure}
At the beginning of the experiment, the participants read and signed a consent form. 
After filling in the pre-experiment questionnaire (Section \ref{pre-questionnaire}), the participants were given a general explanation about the experimental area and the task (including an explanation about teleoperating the robot around an obstacle and treating the patient) without going into great detail, so as not to create bias.
The participants were then asked to move towards the control position as shown in Fig.~\ref{cross-section}.
Each participant experimented with all three LOEs. 
The order of LOEs for each individual participant was chosen by a balanced Latin square design.
After completion of the experiment, participants were required to fill in a final questionnaire (Section \ref{final-questionnaire}). 
Thereafter, they were interviewed by the experimenter, who asked the three questions described in Section \ref{Interview}.

%All experiments were approved by the department ethics board.
%\subsection{Measures}
%Independent variables were LOEs
\subsection{Measures and interviews with the participants}
\subsubsection{\textbf{Pre-experiment questionnaire}}
\label{pre-questionnaire}
The pre-experiment questionnaire comprised questions regarding demographic information (e.g., age, gender) together with the Negative Attitude toward Robots Scale (NARS) questionnaire~\cite{syrdal2009negative} to assess – on a 5-point Likert-type scale ranging from 1 (``Strongly disagree") to 5 (``Strongly agree") – whether the participant had a negative attitude toward situations involving interaction with robots.

\subsubsection{\textbf{Subjective Measures}} \label{final-questionnaire}
Subjective measures were assessed via the final questionnaire, shown in Table~\ref{tab:FinalQ}, using a 5-point Likert-type scale ranging from 1 (``Strongly disagree") to 5 (``Strongly agree") for all measures. %Each question in the questionnaire was taken from previous work in the HRI community, as illustrated in Appendix \ref{FinalQ}.
The final questionnaire included three subjective measures, Fluency of interaction~\cite{8678448}, Adequacy of explanation~\cite{hoffman2018metrics}, and Trust~\cite{hoffman2018metrics}, which the participants rated for each of the different LOEs.
\begin{table*}[h!]
\caption{Subjective measures - Questions by measures as taken from previous work in the HRI community}
\label{tab:FinalQ}
\resizebox{\columnwidth}{!}{%
\begin{tabular}{|c|c|c|}
\hline
\textbf{Measure}         & \textbf{Question}           & \textbf{Reference}       \\ \hline
\multirow{2}{*}{\textbf{\begin{tabular}[c]{@{}c@{}}Human-oriented -\\ Fluency of interaction\end{tabular}}} & I felt the information provided by the robot was given at the right time.                                                                 & \multirow{2}{*}{\cite{8678448}} \\ 
                        & I felt like the robot was committed to success. & \\ \hline
\multirow{3}{*}{\textbf{\begin{tabular}[c]{@{}c@{}}Adequacy of \\    explanation\end{tabular}}}             & \begin{tabular}[c]{@{}c@{}}The robot’s explanations were sufficient and sufficiently \\ detailed for me to do my task.\end{tabular} & \multirow{3}{*}{\cite{hoffman2018metrics}} \\ 
                                & The robot’s explanations were satisfactory.       &  \\ 
                                & The explanation is actionable, and I felt the robot and I were a good team.&     \\ \hline
\multirow{3}{*}{\textbf{Trust}}                      & I felt stressed, worried or had doubts when the robot gave instructions.                                                          & \multirow{3}{*}{\cite{hoffman2018metrics}} \\
        & I had confidence in the robot’s instructions.        &               \\ 
            & My trust in the robot and its abilities was high.    &   \\ \hline
\end{tabular}%
}
\end{table*}

\subsubsection{Objective Measures}
The following four objective measures were derived for each LOE (Table~\ref{tab:objective}):  the time to bypass the obstacle, the number of collisions that happened due to the robot teleoperation by the user, the number of incorrect movements executed by the robot during teleoperation, and the number of times participants asked for clarification. 
%% Start of new table 
\begin{table}{}\
\caption{Different objective measures taken from the experiment.}
\label{tab:objective}
\begin{tabular}{|l|l|}\hline
Objective Measure                                                  & Explanation                                                                                                                                                                                                                                                                                                                         \\ \hline
\begin{tabular}[c]{@{}l@{}}Completion\\ time\end{tabular}           & \begin{tabular}[c]{@{}l@{}}The time (in seconds) that the robot took from \\ the moment that the user received the first\\ explanation for bypassing an obstacle  until s/he \\ bypassed it, plus the time it took him/her to give the \\ correct treatment to the patient from the \\ moment that the robot arrived at the patient.\end{tabular} \\ \hline
\begin{tabular}[c]{@{}l@{}}Number\\  of collisions\end{tabular}     & \begin{tabular}[c]{@{}l@{}}The number of times the robot collided with \\ an obstacle when the human was controlling it.\end{tabular}                                                                                                                                                                                                \\ \hline
\begin{tabular}[c]{@{}l@{}}Incorrect \\ Movements\end{tabular}          & \begin{tabular}[c]{@{}l@{}}The number of times the user did not follow\\  the explanations provided and navigated the\\ robot in a different path.\end{tabular}                                                                                                                                                                  \\ \hline
\begin{tabular}[c]{@{}l@{}}Number of \\ clarifications\end{tabular} & \begin{tabular}[c]{@{}l@{}}The number of times the user sought \\ help from the experimenter.\end{tabular}                                                                                                                                                                                                                       \\ \hline
\end{tabular}

\end{table}
\subsubsection{\textbf{Interview}} \label{Interview}
The interview with the participants after completing the experiment was based on the following three questions, which were similar for the two task conditions (with and without a time limit):

\begin{enumerate}
    \item At which obstacle was the explanation the best for you? Why?
    \item  What were your feelings during the experiment at each of the obstacles? 
    \item If there was no time limit (for the \textit{`with time limit’} task condition, and vice versa for the \textit{`without time limit’} task condition), do you think that a different level of explanation would have been more appropriate for the task? 
\end{enumerate} 

\subsection{Analysis}
%Descriptive statistics (mean ± standard deviation for normal distribution, median for non-normal distribution, and percentage for ordinal data) were computed for each of the dependent variables. 
%Dependent variables that consisted only of subjective sub-variables were averaged, and variables that were composed of both subjective and objective sub-variables were first normalized between 0-1 and then averaged. 
The NARS questionnaire contains three categories (negative attitudes toward situations of interaction with robots, negative attitudes toward the social influence of robots, and negative attitudes toward emotions in interaction with robots) with multiple questions for each category. 
The responses of each participant were averaged for each category.
Similarly, for each subjective measure (fluency of interaction, adequacy of explanation, and trust), there was more than one question (see Table~\ref{tab:FinalQ}).
We took the mean of the participant's responses for the metrics. 

%ORDER EFFFECT WAS ANALYZED HOW.

The Kolmogorov-Smirnov test was conducted on each dependent variable to check for normal distribution. The analysis revealed that most of the variables (except NARS questionnaire and completion time) were not normal.
Therefore, the Friedman test was conducted on each dependent variable.

We conducted repeated measures analysis of variance (ANOVA) for the completion time.
We report $\chi^2$ for Friedman test and $F-value$ for ANOVA, p-values ($p$) and effect size ($r$).
%The analysis revealed that \textit{completion time} (D = 0.118, p = 0.013) and \textit{fluency of interaction} ($D = 0.239, p <0.001$) were not normally distributed, but the \textit{adequacy of explanation} (D = 0.058, p = 0.762) and \textit{trust} (D = 0.086. p = 0.327) were normally distributed.

For all independent variables (LOE) that had a significant effect  on the dependent variable,  a posthoc test (pair-wise Wilcoxon-test with Bonferroni correction and post-hoc Tukey for completion time) was conducted using least squares means.
%In the interaction variables, we examined whether the independent variables had an effect on each other.
%Welch’s T-test for independent samples was applied to compare between \textit{`with time limit'} and \textit{`without time limit'} conditions in terms of \textit{adequacy of explanation} and \textit{trust}, and the Mann-Whitney U test was applied to compare between \textit{`with'} and \textit{`without time limit’} in terms of \textit{completion time} and \textit{fluency of interaction}.

To ensure that there were no contradictions in the participants’ answers and that the questions did indeed examine the same measure, we examined the correlations between the questions in the final questionnaire that related to the same variable. 

The interviews that were conducted with participants were recorded and then analyzed. The first step in the analysis was to determine the most popular four responses to each question. Then, for each question, all answers of all partIcipants were classified according to these four common responses (see section~\ref{subsec:inter_withlimit} and section~\ref{subsec:inter_withouttime}).

\section{Results}
\subsection{Pre-Experiment Questionnaire}
The NARS questionnaire is divided into three subcategories, negative attitudes toward situations of interaction with robots, negative attitudes toward the social influence of robots, and negative attitudes toward emotions in interaction with robots.
A t-test was conducted to compare the groups' negative attitudes toward the robot.
It was found that for participants in both groups, there was no significant difference between the two groups of participants in negative attitudes toward situations of interaction with robots (t(58) = 1.08, p = 0.282), negative attitudes toward the social influence of robots (t(58) = -1.12, p = 0.267), and negative attitudes toward emotions in interaction with robots (t(58) = -1.55, p = 0.126).
%The statistical comparison between the LOEs (high, medium, and low) for each dependent variable and condition (\textit{`with time limit'} and \textit{`without time limit'}) is given in Table \ref{tab:ZTP}. The z.ratio (for a non-normal distribution), the t.ratio (for a normal distribution), and the p-value are reported. Significant and non-significant variables are indicated in green and pink, respectively.
\subsection{Final Questionnaire}
Results revealed that there was a significant correlation between all the questions relating to the same variable [fluency of interaction (two questions): correlation = 0.357, p = 0.048; adequacy of explanation (three questions): correlations = 0.77, 0.64, 0.71, $p < 0.001$; trust (three questions): correlations = 0.33, 0.53, 0.5, $p = 0.01$, $<0.001$, $<0.001$]. All the tests were designed as two-tailed with a significance level of 0.05.

\subsection{\textit{`With time limit'}}
{The comparison tests (Friedman or ANOVA) were conducted on the subjective measures (see Table~\ref{tab:FinalQ}) and objective measures (see Table~\ref{tab:objective}).
The responses to the interview questions are detailed in section~\ref{subsec:inter_withlimit} and section~\ref{subsec:inter_withouttime}.
\subsubsection{Metrics}
For both objective and subjective measures, there was a significant difference between the LOE when either Friedman's test or ANOVA was conducted:  fluency of interaction ($\chi^2 (2)= 31.764, p<0.001,r=0.523$), 
adequacy of explanations ($\chi^2 (2)= 34.43, p<0.001,r=0.57$), 
trust ($\chi^2 (2)= 33.85, p<0.001, r=0.56$), 
number of collisions ($\chi^2 (2)= 43.57, p<0.001,r=0.726$), incorrect movements ($\chi^2 (2)= 54.44, p<0.001,r=0.91$), 
number of clarifications ($\chi^2 (2)= 36.20, p<0.001,r=0.60$)
and completion time ($F(2,87)=5.23,p=0.0072, r=0.11$).

Post-hoc tests revealed that there were significant differences between high and low LOE and between medium and low LOE for both subjective measures (see Table~\ref{tab:ZTP}) and objective measures (see Table~\ref{tab:post_objective}).
There was no difference between high and medium LOE for subjective  (see Table~\ref{tab:ZTP}) and objective measures (see Table~\ref{tab:post_objective}).
High and medium LOE were preferred over low LOE for all subjective measures, as shown in Fig.~\ref{fig:sub_with_time}.
In the low LOE, participants took more time to complete the task (see Fig.~\ref{fig:completiontime}), navigated the robot with more collisions with the obstacle (see Fig.~\ref{fig:objective_with_time}), and made more incorrect movements (see Fig.~\ref{fig:objective_with_time}). 
Participants also required more clarifications from the experimenter (see Fig.~\ref{fig:objective_with_time}) for the low LOE as compared to the medium and high LOE.
%%%%%%%%%%%%%%% Start OF TABLE %%%%%%%%%%%%%%%

\begin{table*}[h!]
\caption{Post hoc comparison of subjective measures and completion time. 
Significant (green) and non significant (red) differences between the LOE.
}
\label{tab:ZTP}

\resizebox{1.0\linewidth}{!}{%
\begin{tabular}{|c|cc|cc|cc|cc|}
\hline
\textbf{}                                                                        & \multicolumn{2}{c|}{\textbf{Completion time}}                                                                                                                                                                              & \multicolumn{2}{c|}{\textbf{Fluency of interaction}}                                                                                                                                                                       & \multicolumn{2}{c|}{\textbf{Adequacy of explanation}}                                                                                                                                                                       & \multicolumn{2}{c|}{\textbf{Trust}}                                                                                                                                                                                        \\ \hline
\textbf{LOE}                                                                     & \multicolumn{1}{c|}{\textbf{\textit{\begin{tabular}{c}
     `With\\ time limit'\end{tabular}}}}                                                                               & \textbf{\textit{\begin{tabular}{c}
     `Without \\time limit'  
\end{tabular}}}                                                                            & \multicolumn{1}{c|}{\textbf{\textit{\begin{tabular}{c}
     `With\\ time limit'\end{tabular}}}}                                                                               & \textbf{\textit{\begin{tabular}{c}
     `Without \\time limit'  
\end{tabular}}}                                                                            & \multicolumn{1}{c|}{\textbf{\textit{\begin{tabular}{c}
     `With\\ time limit'\end{tabular}}}}                                                                               & \textbf{\textit{\begin{tabular}{c}
     `Without \\time limit'  
\end{tabular}}}                                                                             & \multicolumn{1}{c|}{\textbf{\textit{\begin{tabular}{c}
     `With\\ time limit'\end{tabular}}}}                                                                               & \textbf{\textit{\begin{tabular}{c}
     `without \\time limit'  
\end{tabular}}}                                                                            \\ \hline
\textbf{\begin{tabular}[c]{@{}c@{}}High\\ vs.\\  Low\end{tabular}}    & \multicolumn{1}{c|}{\cellcolor[HTML]{C5E0B3}\begin{tabular}[c]{@{}c@{}}t.ratio = -2.90\\    $p =0.013$\end{tabular}}  & \cellcolor[HTML]{C5E0B3}\begin{tabular}[c]{@{}c@{}}t.ratio = -4.16\\    $ p < 0.001$\end{tabular}  & \multicolumn{1}{c|}{\cellcolor[HTML]{C5E0B3}\begin{tabular}[c]{@{}c@{}}z.ratio = -4.32\\     $p < 0.001$\end{tabular}}   & \cellcolor[HTML]{C5E0B3}\begin{tabular}[c]{@{}c@{}}z.ratio = -4.14\\   $ p < 0.001$\end{tabular}  & \multicolumn{1}{c|}{\cellcolor[HTML]{C5E0B3}\begin{tabular}[c]{@{}c@{}}z.ratio = -4.63\\    $p < 0.001$\end{tabular}}  & \cellcolor[HTML]{C5E0B3}\begin{tabular}[c]{@{}c@{}}z.ratio = 4.53\\     $p < 0.001$\end{tabular}  & \multicolumn{1}{c|}{\cellcolor[HTML]{C5E0B3}\begin{tabular}[c]{@{}c@{}}a.ratio = -4.45\\    $p < 0.001$\end{tabular}} & \cellcolor[HTML]{C5E0B3}\begin{tabular}[c]{@{}c@{}}z.ratio = -4.47\\     $p < 0.001$\end{tabular}  \\ \hline
\textbf{\begin{tabular}[c]{@{}c@{}}High \\vs.\\ Medium\end{tabular}} & \multicolumn{1}{c|}{\cellcolor[HTML]{FEC4C4}\begin{tabular}[c]{@{}c@{}}t.ratio = -0.213\\     p = 0.98\end{tabular}}   & \cellcolor[HTML]{FEC4C4}\begin{tabular}[c]{@{}c@{}}t.ratio = -1.58\\    p = 0.26\end{tabular}   & \multicolumn{1}{c|}{\cellcolor[HTML]{FEC4C4}\begin{tabular}[c]{@{}c@{}}z.ratio = -1.34\\    p = 0.54\end{tabular}}   & \cellcolor[HTML]{FEC4C4}\begin{tabular}[c]{@{}c@{}}z.ratio = -1.94\\     p = 0.155\end{tabular}  & \multicolumn{1}{c|}{\cellcolor[HTML]{FEC4C4}\begin{tabular}[c]{@{}c@{}}z.ratio = -1.6\\     p = 0.327\end{tabular}}   & \cellcolor[HTML]{C5E0B3}\begin{tabular}[c]{@{}c@{}}z.ratio = -3.39\\    p = 0.002\end{tabular}   & \multicolumn{1}{c|}{\cellcolor[HTML]{FEC4C4}\begin{tabular}[c]{@{}c@{}}z.ratio = -1.83\\    $p = 0.202$\end{tabular}}  & \cellcolor[HTML]{FEC4C4}\begin{tabular}[c]{@{}c@{}}z.ratio = -2.24\\    p = 0.076\end{tabular}  \\ \hline
\textbf{\begin{tabular}[c]{@{}c@{}}Medium\\  vs. \\Low\end{tabular}}  & \multicolumn{1}{c|}{\cellcolor[HTML]{C5E0B3}\begin{tabular}[c]{@{}c@{}}t.ratio = 2.69\\     $p =0.023$\end{tabular}} & \cellcolor[HTML]{C5E0B3}\begin{tabular}[c]{@{}c@{}}t.ratio = -2.58\\     p = 0.03\end{tabular} & \multicolumn{1}{c|}{\cellcolor[HTML]{C5E0B3}\begin{tabular}[c]{@{}c@{}}z.ratio = -4.05\\    $p < 0.001$\end{tabular}} & \cellcolor[HTML]{C5E0B3}\begin{tabular}[c]{@{}c@{}}z.ratio = -4.00\\    $p < 0.001$\end{tabular} & \multicolumn{1}{c|}{\cellcolor[HTML]{C5E0B3}\begin{tabular}[c]{@{}c@{}}z.ratio = -4.17\\    $p < 0.001$\end{tabular}} & \cellcolor[HTML]{C5E0B3}\begin{tabular}[c]{@{}c@{}}z.ratio = -4.21\\     $p < 0.001$\end{tabular} & \multicolumn{1}{c|}{\cellcolor[HTML]{C5E0B3}\begin{tabular}[c]{@{}c@{}}z.ratio = -3.90\\     $p < 0.001$\end{tabular}} & \cellcolor[HTML]{C5E0B3}\begin{tabular}[c]{@{}c@{}}z.ratio = -4.21\\     $p < 0.001$\end{tabular} \\ \hline
\end{tabular}}
\end{table*}
%%%%%%%%%%%%%%% END OF TABLE %%%%%%%%%%%%%%%
\begin{table*}[h!]
\caption{Post hoc comparison of objective measures 
Significant (green) and non significant (red) differences between the LOE.
}
\label{tab:post_objective}

\resizebox{\linewidth}{!}{

\begin{tabular}{|l|ll|ll|ll|}

\hline                                                                & \multicolumn{2}{l|}{\textbf{Number   of collisions}}                            & \multicolumn{2}{l|}{\textbf{Incorrect Movements}}                                  & \multicolumn{2}{l|}{\textbf{Number   of clarifications}}                                                                                                                                                                                                                     \\ \hline
\textbf{LOE}                                                            & \multicolumn{1}{l|}{\textbf{\begin{tabular}{c}
     `With \\time limit’
\end{tabular}}}                                                                                                   & \textbf{\begin{tabular}{c}
     `Without \\ time limit’
\end{tabular}}                                                                                                  & \multicolumn{1}{l|}{\textbf{\begin{tabular}{c}
     `With \\time limit’
\end{tabular}}}                                                                                                     & \textbf{\begin{tabular}{c}
     `Without \\ time limit’
\end{tabular}}                                                                                                  & \multicolumn{1}{l|} {\textbf{\begin{tabular}{c}
     `With \\time limit’
\end{tabular}}}                                                                                                     & {\textbf{\begin{tabular}{c}
     `Without \\ time limit’
\end{tabular}}}                                                                                                  \\ \hline
\textbf{\begin{tabular}[c]{@{}l@{}}High\\ vs.\\ Low\end{tabular}}    & \multicolumn{1}{l|}{\cellcolor[HTML]{C5DFB3}\begin{tabular}[c]{@{}l@{}}z.ratio = -4.44   \\ $p < 0.001$\end{tabular}} & \cellcolor[HTML]{C5DFB3}\begin{tabular}[c]{@{}l@{}}z.ratio = -4.25\\    $p <0.001$\end{tabular} & \multicolumn{1}{l|}{\cellcolor[HTML]{C5DFB3}\begin{tabular}[c]{@{}l@{}}z.ratio = -4.80\\    p \textless   0.001\end{tabular}} & \cellcolor[HTML]{C5DFB3}\begin{tabular}[c]{@{}l@{}}z.ratio = -4.81\\    p \textless   0.001\end{tabular} & \multicolumn{1}{l|}{\cellcolor[HTML]{C5DFB3}\begin{tabular}[c]{@{}l@{}}z.ratio = -4.08\\    p \textless   0.001\end{tabular}} & \cellcolor[HTML]{C5DFB3}\begin{tabular}[c]{@{}l@{}}z.ratio = -4.16\\    p \textless   0.001\end{tabular} \\ \hline
\textbf{\begin{tabular}[c]{@{}l@{}}High \\vs.\\    Medium\end{tabular}} & \multicolumn{1}{l|}{\cellcolor[HTML]{FEC4C4}\begin{tabular}[c]{@{}l@{}}z.ratio = -0.94\\    p = 1.0\end{tabular}}           & \cellcolor[HTML]{C5DFB3}\begin{tabular}[c]{@{}l@{}}z.ratio = -2.65\\    p = 0.025\end{tabular}           & \multicolumn{1}{l|}{\cellcolor[HTML]{FEC4C4}\begin{tabular}[c]{@{}l@{}}z.ratio = -2.09\\    p = 0.11\end{tabular}}            & \cellcolor[HTML]{C5E0B3}\begin{tabular}[c]{@{}l@{}}z.ratio =   -3.31\\  p = 0.003\end{tabular}         & \multicolumn{1}{l|}{\cellcolor[HTML]{FEC4C4}\begin{tabular}[c]{@{}l@{}}z.ratio = -2.04\\    p = 0.123\end{tabular}}           & \cellcolor[HTML]{C5DFB3}\begin{tabular}[c]{@{}l@{}}z.ratio = -2.43\\     p = 0.044\end{tabular}           \\ \hline
\textbf{\begin{tabular}[c]{@{}l@{}}Medium \\vs.\\   Low\end{tabular}}  & \multicolumn{1}{l|}{\cellcolor[HTML]{C5DFB3}\begin{tabular}[c]{@{}l@{}}z.ratio = -4.33\\    p \textless 0.001\end{tabular}} & \cellcolor[HTML]{C5DFB3}\begin{tabular}[c]{@{}l@{}}z.ratio = -3.36\\    p = 0.002\end{tabular}           & \multicolumn{1}{l|}{\cellcolor[HTML]{C5DFB3}\begin{tabular}[c]{@{}l@{}}z.ratio = -4.80\\   p \textless   0.001\end{tabular}} & \cellcolor[HTML]{C5DFB3}\begin{tabular}[c]{@{}l@{}}z.ratio = -4.64\\   p \textless   0.001\end{tabular} & \multicolumn{1}{l|}{\cellcolor[HTML]{C5DFB3}\begin{tabular}[c]{@{}l@{}}z.ratio = -3.90\\     p \textless   0.001\end{tabular}} & \cellcolor[HTML]{C5DFB3}\begin{tabular}[c]{@{}l@{}}z.ratio = -3.55\\    p = 0.001\end{tabular}           \\ \hline
\end{tabular}}
\end{table*}
\begin{figure*}[h!]
    \centering
    
    \subfloat[\textit{With time limit}]{\includegraphics[width=\linewidth]{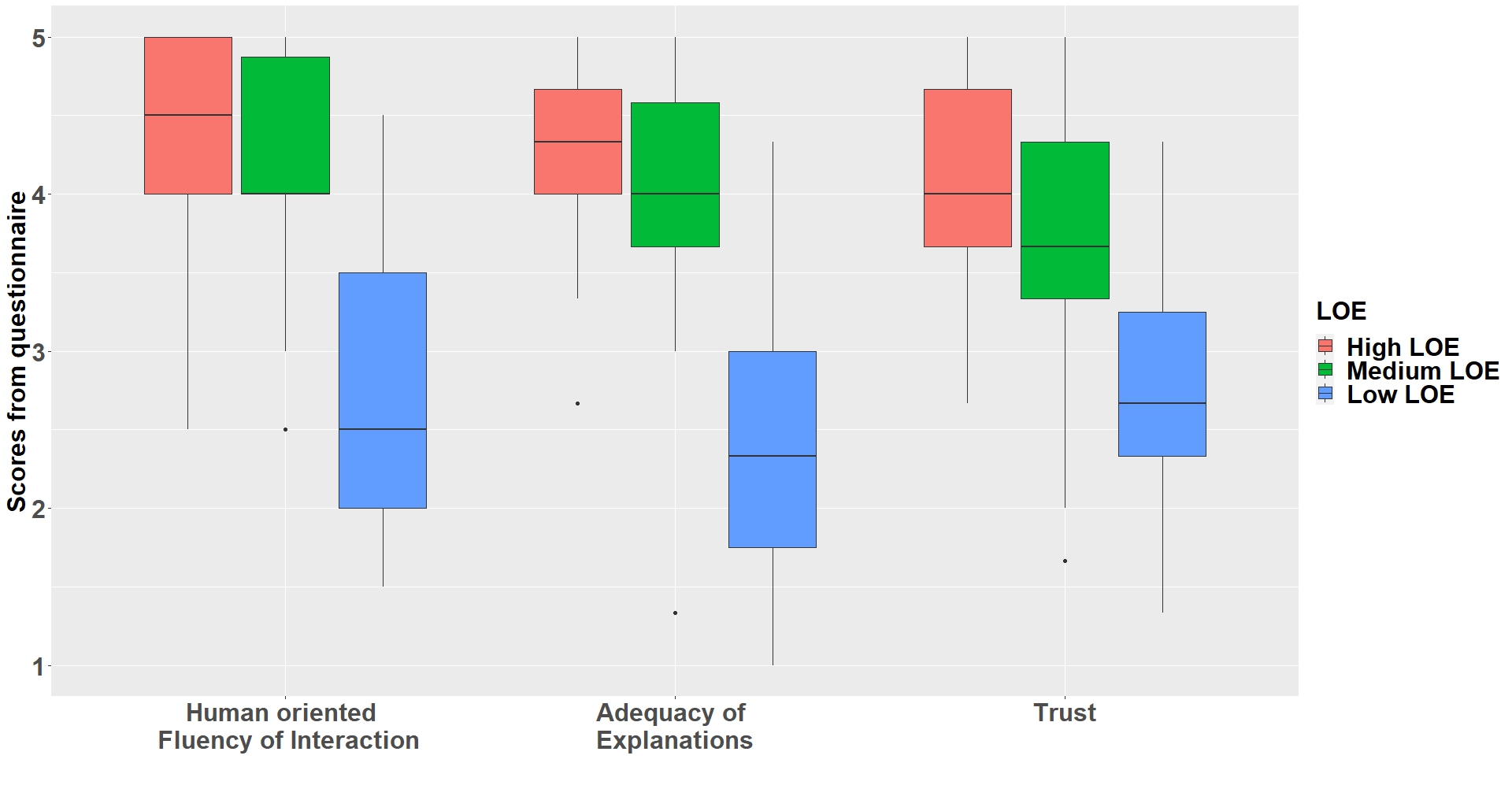}
    \label{fig:sub_with_time}}
    \hspace{0.01cm}
    \subfloat[\textit{Without time limit}]{\includegraphics[width=\linewidth]{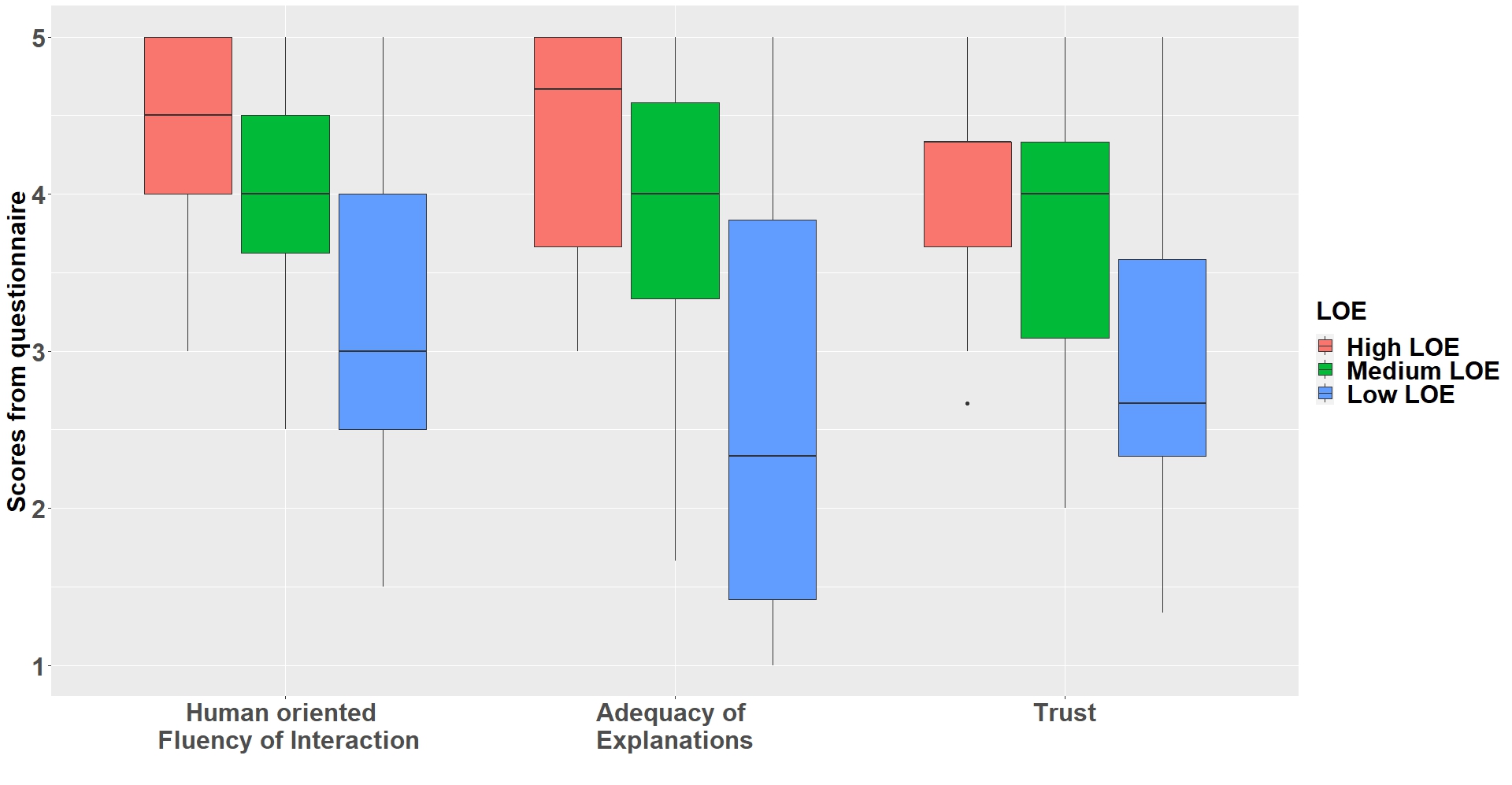}
    \label{fig:sub_without_time}}
    \caption{ Boxplots for subjective measures for all the three levels in both conditions (a) i.e., \textit{with time limit} and (b)\textit{without time limit} condition. }
    \label{fig:subjective_measure}
\end{figure*}
%%%  style for figure starts here
%
\begin{figure*}
    \centering
    \includegraphics[width=\linewidth]{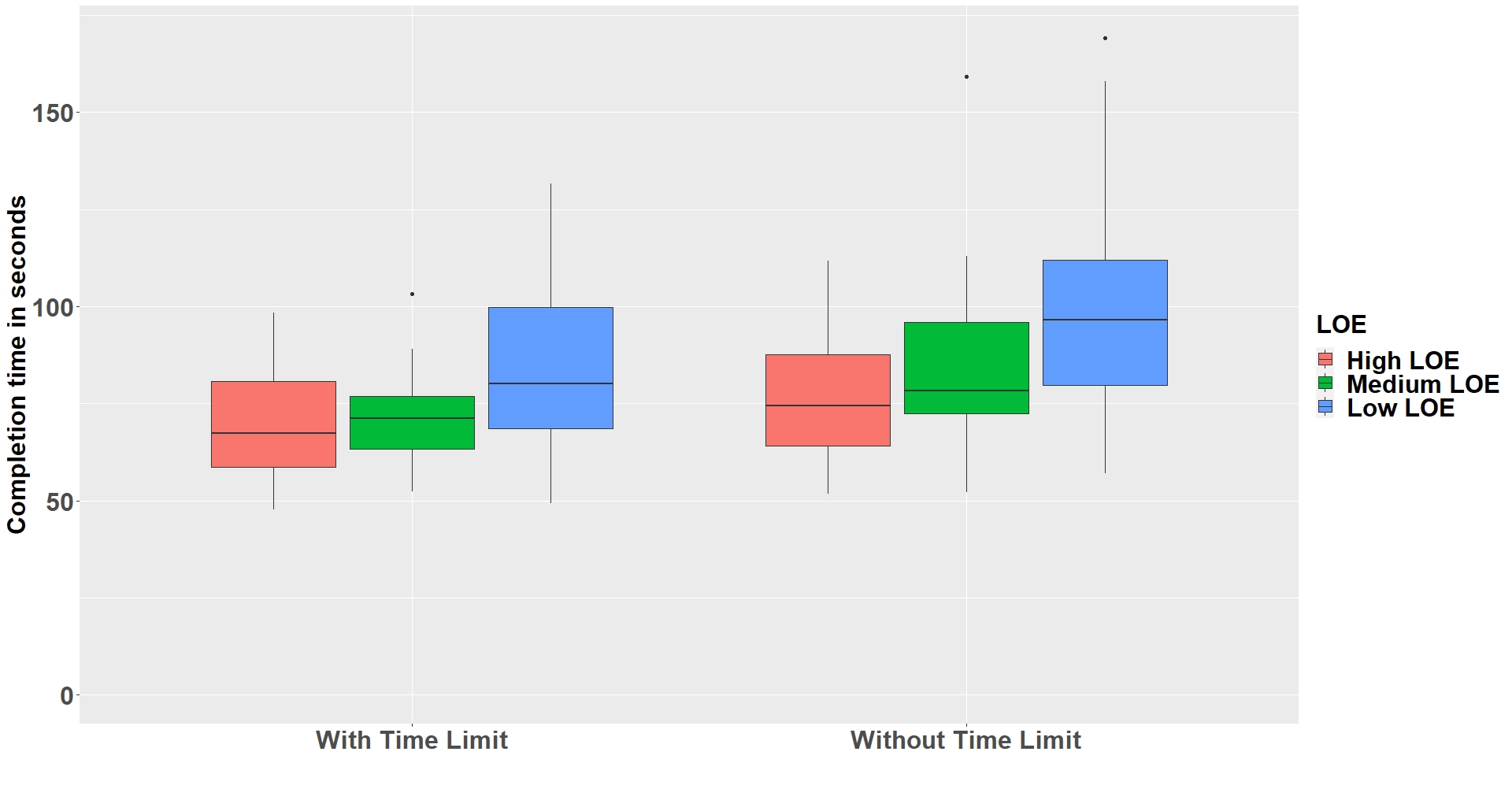}
    \caption{ Boxplot for completion for all the three levels in both conditions i.e., \textit{with time limit} and \textit{without time limit} condition.}
    \label{fig:completiontime}
\end{figure*}

\begin{figure*}[h!]
\centering
\subfloat[\textit{With time limit}]
{\includegraphics[width=\linewidth]{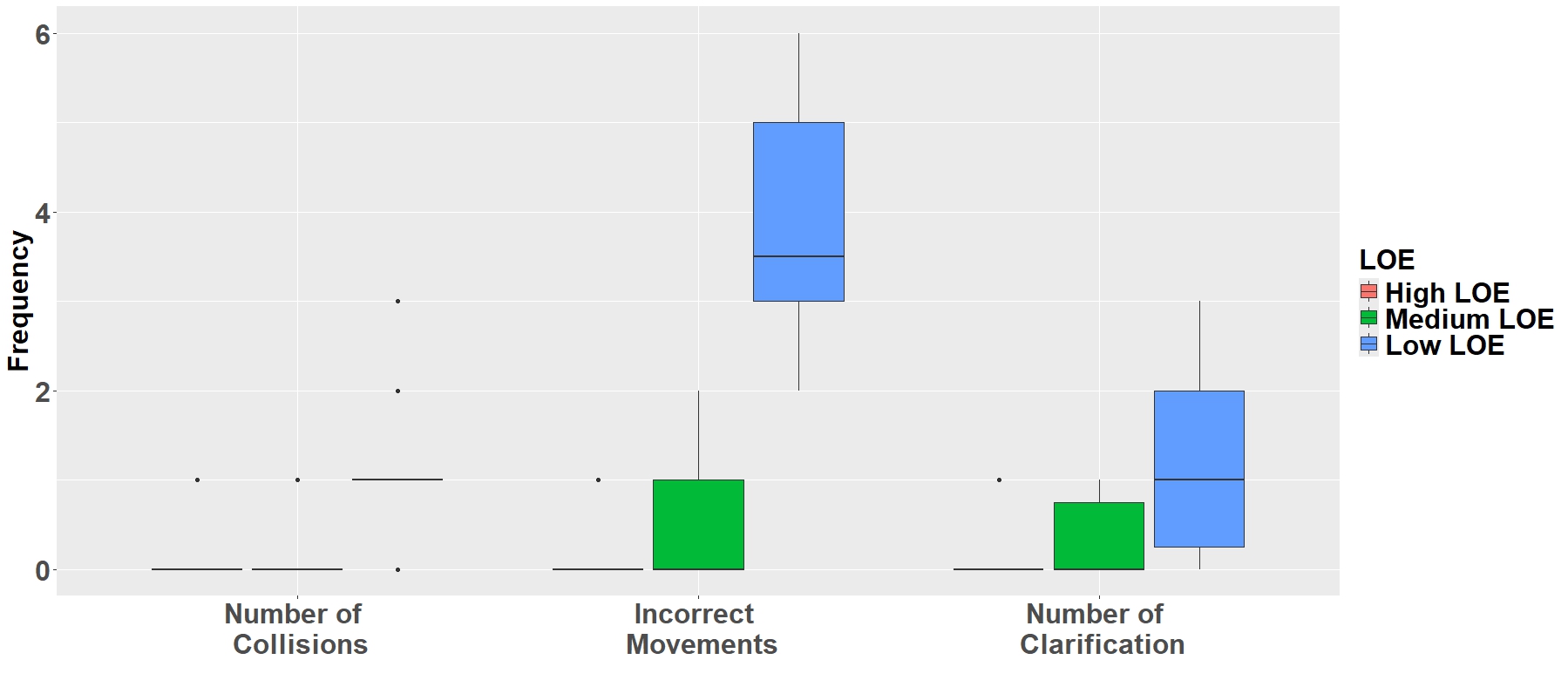}
\label{fig:objective_with_time}}
\hspace{0.1cm}
\subfloat[\textit{Without time limit}]
{\includegraphics[width=\linewidth]{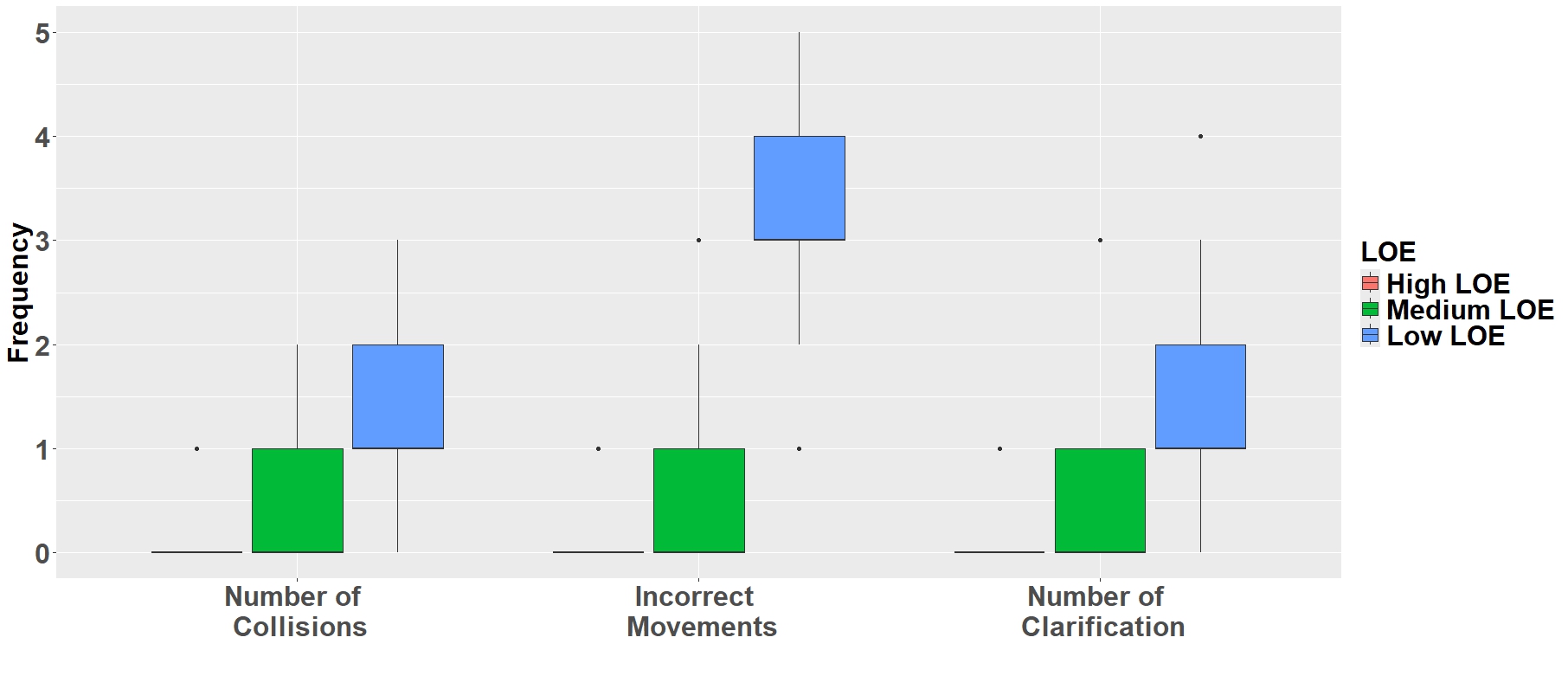}
\label{fig:objective_without_time}}
\caption{Boxplots of objective measures (number of collisions, incorrect movements, and number of clarifications) for all LOEs in (a) \textit{with time limit} (b)\textit{without time limit} condition.}
\end{figure*}
\subsubsection{Interviews}\label{subsec:inter_withlimit}
 Of the participants, 43.3\% stated that they preferred the high LOE because the explanation was very detailed, helped them operate the robot correctly under time pressure, and hence made them feel comfortable (Table~\ref{tab:InterviewTableApendix}). A similar number, i.e., 40\%, stated that the medium LOE was the preferred level under the time limit pressure because the explanation was sufficient for them to understand what needed to be done without being too detailed, which would have caused delays in the execution of the task (Table~\ref{tab:InterviewTableApendix}).
Only 6.67\% of the participants stated that they preferred the low LOE under time pressure; 
those participants believed that they could perform the task with only a general explanation because they trusted themselves more than the robot’s explanations. In addition, 10\% of the participants stated that they had no preference between high and low LOE (Table~\ref{tab:InterviewTableApendix}).

When the participants were asked about their feelings regarding each of the LOE, 93.33\% stated that at the high and medium LOE they felt that the explanations were clear and enabled them to perform the task well. 
In contrast, they felt that the low LOE was burdensome and confusing and led to a lack of understanding on their part. The remaining 6.67\% of participants stated that they felt that the low LOE was the most suitable level under a time limit because they received a general explanation and could then act quickly and precisely, whereas the other LOE would hinder their level of competence. When asked about which LOE they would prefer if there was no time limit, 43.3\% of the participants stated that they would prefer the high LOE, 33.3\%, the medium LOE, and only 3.3\%, the low LOE. In addition, 20\% stated that they would prefer the high and medium LOE equally.

\subsection{\textbf{\textit{`Without time limit'} condition}}\label{ subsec:withot}
Similar to the previous section, we report a comparison test of subjective measures (Table~\ref{tab:FinalQ}) and objective measures (Table~\ref{tab:objective}) in \textit{`without time limit'}. We also report the interview responses. 
\subsubsection{Metrics}
All the dependent variables were significantly different among all the three LOE, i.e., fluency of interaction ($\chi^2 (2)= 34.44, p<0.001,r=0.57$), adequacy of explanations ($\chi^2 (2)= 42.56, p<0.001,r=0.53$), 
trust ($\chi^2 (2)= 37.82, p<0.001,r=0.63$), 
number of collisions ($\chi^2 (2)= 31.83, p<0.001,r=0.73$), 
wrong movements ($\chi^2 (2)= 53.84, p<0.001,r=0.90$), 
number of clarification ($\chi^2 (2)= 30.48, p<0.001,r=0.51$) and completion time ($F(2,87)=8.83,p<0.001,r=0.17$). 

Participants found high LOE significantly different (Table~\ref{tab:ZTP}) and 
adequately explainable (Fig.~\ref{fig:sub_without_time}) vs. medium LOE and low LOE.
Similarly medium LOE was significantly different (see Table~\ref{tab:ZTP}) and adequately explainable (Fig.~\ref{fig:sub_without_time}) vs. low LOE.
For completion time, fluency of interaction, and trust, there was a significant difference between high and low LOE as well as between medium and low LOE, as shown in Table~\ref{tab:ZTP}.
However, there was no significant difference between high and medium LOE as shown in Table~\ref{tab:ZTP}.
For high LOE and medium LOE, participants took less time 
(Fig.~\ref{fig:completiontime}) and perceived interaction as more fluent and trustworthy compared to low LOE, as shown in Table~\ref{tab:ZTP}.

For all objective measures (Table~\ref{tab:post_objective}; 
number of collisions, wrong movements, and number of clarifications), there were significant differences between high versus low, medium versus low, and high versus medium LOE. 
Participants navigated the robot with 
fewer collisions (Fig.~\ref{fig:objective_without_time}), and fewer incorrect movements, and required fewer clarifications in high LOE compared to low and medium LOE.
Similarly, participants performed better in medium LOE as compared to low LOE.

\subsubsection{Interviews}\label{subsec:inter_withouttime}
Of the participants, 53.3\%  stated that they preferred the high LOE because the explanation was very detailed, helped them operate the robot correctly under time pressure, and made them feel comfortable (Table~\ref{tab:InterviewTableApendix}). 
They claimed that without a time limit, they would like to receive as many details as possible and also to receive an explanation that was as reasoned as possible. In contrast, 30\% stated that the medium LOE was the preferred level because it was sufficiently understandable and they did not feel the need for details, even if there was no time limit (Table~\ref{tab:InterviewTableApendix}).
Of the remaining participants, only 3.3\% stated that they would prefer the low LOE, and 13.3\% stated that they had no preference between the high and low LOE (Table~\ref{tab:InterviewTableApendix}).

In summary, when the participants were asked about their feelings regarding each of the LOE, 96.6\% stated that for both the high and medium LOE they felt that the explanations were clear and allowed them to perform the task well. In contrast, they felt that the low LOE was burdensome and confusing and led to a lack of understanding. Only 3.3\% of the participants stated that they felt that the low LOE was the most suitable level because they preferred to act more independently rather than to listen to explanations.

When asked which LOE they would prefer if they had a time limit, 43.3\%, 40\%, and 3.3\% of the participants stated that they would prefer the high, medium, and low LOE, respectively, and 13.3\% stated that they had no preference regarding the high vs. the medium LOE.

\begin{table*}[h!]
\caption{Classification of answers to each question from the interview analyses. }
\label{tab:InterviewTableApendix}
\resizebox{\columnwidth}{!}{%
\begin{tabular}{|c|c|l|l|}
\hline
\multirow{8}{*}{\begin{tabular}[c]{@{}c@{}}Both conditions\\      (\textit{`with time limit'}, \textit{`without time limit'})\end{tabular}} & \multirow{4}{*}{Question 1} & 1 & \begin{tabular}[c]{@{}l@{}}In the high LOE, the explanation was the best for me\\ because it was the most detailed and understandable.\end{tabular}  \\ \cmidrule{3-4}  &  & 2 & \begin{tabular}[c]{@{}l@{}}In the medium LOE, the explanation was the best for me \\ because the explanation in it was just at the right amount, not too much, like the high level,\\ and not missing, like the low level.\end{tabular} \\ \cmidrule{3-4}  &  & 3 & \begin{tabular}[c]{@{}l@{}}In the low LOE, the explanation was the best for me \\ because I prefer to receive a general explanation and then trust my abilities\\ without having to use explanations from the robot.\end{tabular}  \\ \cmidrule{3-4}&  & 4 & \begin{tabular}[c]{@{}l@{}}I felt that the explanation in both high and medium levels of explanation\\ was equally good for me and better than the low level.\end{tabular} \\ \cmidrule{2-4}  & \multirow{4}{*}{Question 2} & 1 & \begin{tabular}[c]{@{}l@{}}At the high level, I felt the most comfortable and confident operating the robot, \\ at the medium level, I also felt relatively good but it was less well understood, \\ but at the low level, I really felt that I did not understand the robot's explanations\\ and I didn't know what to do, so I had to rely more on myself and I used only\\ the camera and I ignored the instructions which confused me.\end{tabular} \\ \cmidrule{3-4} &                             & 2 & \begin{tabular}[c]{@{}l@{}}At the medium level, I felt the most comfortable and confident operating the robot, \\ at the high level, I also felt relatively good but it was too detailed and I don't think \\ it's necessary, but at the low level I really felt that I did not understand the robot's \\ explanations and I didn't know what to do, so I had to rely more on myself and \\ I used only the camera and I ignored the explanations which confused me.\end{tabular} \\ \cmidrule{3-4}  &                             & 3 & \begin{tabular}[c]{@{}l@{}}I prefer to trust myself more than explanations from the robot, so I liked the low level\\ of explanation the most compared to the others, even though they are detailed and better\\  in terms of explanation.\end{tabular}  \\  \cmidrule{3-4}   &                             & 4 & \begin{tabular}[c]{@{}l@{}}I felt good and comfortable operating the robot both at the high and medium levels of \\ explanation. I think the explanations at both levels are equally good and better than \\ the low level, which was not clear, made me confused and I didn’t know what to do so \\ I had to perform the task without relying on the robot.\end{tabular}                                                                                                                  \\ \hline
\multirow{4}{*}{\textit{`with time limit'} condition}                                                                       & \multirow{4}{*}{Question 3} & 1 & \begin{tabular}[c]{@{}l@{}}I would choose the high LOE if there ware no time limit for the task - \\ without a time limit I would prefer to get the most detailed explanations I can get.\end{tabular}  \\ \cmidrule{3-4}   &                             & 2 & \begin{tabular}[c]{@{}l@{}}I would choose the medium LOE if there were no time limit for the task - \\ I think it is clear enough and there is no point in getting overly detailed explanations.\end{tabular}                      \\  \cmidrule{3-4} &                             & 3 & \begin{tabular}[c]{@{}l@{}}I would choose the low LOE if there was no time limit - \\ I like to trust myself more than explanations from the robot and if I'm not \\ pressed for time I'd prefer a very general explanation and then operate alone.\end{tabular}             \\ \cmidrule{3-4} &                             & 4 & \begin{tabular}[c]{@{}l@{}}I would choose the high or medium levels of explanation if there was no time limit - \\ I think these two levels might be equally suitable.\end{tabular}    \\ \hline
\multirow{4}{*}{\textit{`without time limit'} condition}                                                                    & \multirow{4}{*}{Question 3} & 1 & \begin{tabular}[c]{@{}l@{}}I would choose the high LOE even if there was a time limit for the task - when \\ I'm under pressure I prefer to receive an exact, clear and detailed explanation so that I can carry out\\ the instructions without thinking too much. It takes the load off and doesn't add pressure like \\ not detailed explanations and will allow me to do the task better.\end{tabular}                                                               \\  \cmidrule{3-4}  &                             & 2 & \begin{tabular}[c]{@{}l@{}}I would prefer the medium LOE in a time limit situation because I want to be on time\\ and the high level is too detailed and may hold me back and in contrast, \\ the low level is not detailed at all and this can only cause me stress and confusion in such a situation.\end{tabular}                                                                                                                                                    \\   \cmidrule{3-4} &                             & 3 & \begin{tabular}[c]{@{}l@{}}I would prefer the low LOE because in a time limit situation I prefer to be in full \\ control of the situation and not be helped by explanations from the robot. \\ When I'm under pressure I prefer to rely only on myself.\end{tabular}                                                                                                                                                                                                   \\ \cmidrule{3-4}    &                             & 4 & \begin{tabular}[c]{@{}l@{}}I would choose the high or medium levels of explanation if there was a time limit - \\ I think these two levels might be equally suitable.\end{tabular}                                                                                                                                                                                                                                                                                                      \\ \hline
\end{tabular}%
}
\end{table*}

\section{Discussion}
The results revealed that the LOE had a significant effect on all the dependent variables and that participants did find differences between the three LOE. 

For all the dependent variables, there was no difference between the high and medium LOE, implying that \textit{H1.1} and \textit{H1.2} were partially validated in the \textit{`with time limit'} condition. 
However, there was a significant difference between the high/medium LOE and the low LOE.

In the \textit{`without time limit'} condition, \textit{H2.1} and \textit{H2.2} were partially validated. 
In the highest LOE, a smaller number of collisions, fewer incorrect movements, and fewer clarifications were required by the participants to complete the task, and the adequacy of explanation was judged to be better. 
For the medium and high LOE, there were no differences in the completion time, fluency of interaction, and trust, but there were significant differences between the high/medium LOE and the low LOE.
In general, irrespective of whether or not there was a time limitation, the high LOE was preferred.

\subsection{\textit{`With time limit'} condition}
Our study showed that in the case of \textit{`with time limit'} either the high or the medium LOE should be provided to the user.
Participants did not differentiate between detailed explanations or brief explanations and their performance (completion time, number of collisions, number of incorrect movements, and number of clarifications)  were not different between medium and high LOE (see Fig.~\ref{fig:objective_with_time} and Fig.~\ref{fig:completiontime}).
This observation stands in contradiction to previous research findings~\cite{7451740}, indicating that only a brief explanation should be provided to facilitate the completion of a time-sensitive task (e.g, search-and-rescue).
However, we stress that our study was conducted in a laboratory environment with engineering students and that the only limitation that our experiment included was a time limit. 
The other measures did not have any effect when a time-sensitivity aspect was introduced in the design.

It has previously been shown that fluency of interaction (see Fig.~\ref{fig:sub_with_time}) improves when the collaborating agent can anticipate the action of the other agent~\cite{hoffman2007effects}. In this study, we devised an explanation intended to help users understand the planning of the robot. We found that the pattern of explanation played an important role in the users' decision regarding the fluency of interaction: The participants felt the interaction to be less fluent in the case of a static pattern of explanation. Since the medium and high LOE were equally preferred between the different conditions, we would suggest that future robotic designers take explanation patterns into consideration when seeking to improve the fluency of interaction.
The adequacy of explanation is defined as clear and precise information provided by the robot~\cite{hoffman2018metrics}.
The level of verbosity (high or low) did not play any role when there was a time limit. This finding suggests that participants receiving only a brief explanation in the case of \textit{`with time limit'} would be able to complete the task (Fig.~\ref{fig:sub_with_time}). 
%Hence, future designers should keep in mind that the robot's explanation should be adaptive according to the demands of the situation.%The current study showed that the user would prefer a detailed explanation in \textit{`without time limit'}. This concurs with our result that a high level of verbosity with a dynamic explanation pattern was the LOE of choice.
A trustor can rely on the trustee only if her/his expectation is matched by the action performed by the trustee~\cite{gurtman1992trust,10.2307/258792}.
The explanation provided by the robot should increase the trust of the user in it~\cite{gunning2019xai}.
Indeed, in our study, high or medium LOE invoked greater trust than low LOE (Fig.~\ref{fig:sub_with_time}).
The users trusted the dynamic explanation pattern to a greater extent than the static pattern. The high vs. low level of verbosity did not have any effect on the trust of the participants, which suggests that participants expected a more dynamic pattern of explanation from the robot in both time conditions.

\subsection{\textit{`Without time limit'} condition}
In a previous study~\cite{gunning2019xai}, it was found that performance could improve with appropriate explanations in the case of no time limit.
We found that the participants completed the task in a shorter time (Fig.~\ref{fig:completiontime}) in the high/medium LOE when there was no time limit as compared to low LOE.
Further, there were fewer collisions and fewer wrong movements, and fewer clarifications (see Fig.~\ref{fig:objective_without_time}) were required in high LOE compared to medium and low LOE.
This demonstrates the importance of having both high verbosity and dynamic patterns to improve human performance during interaction.
%The \textit{adequacy of explanation} is defined as clear and precise information provided by the robot~\cite{hoffman2018metrics}.
The current study showed that the user would prefer a detailed explanation in \textit{`without time limit'} with regards to the adequacy of explanations (see Fig.~\ref{fig:sub_without_time}).  
This finding concurs with our result that a high level of verbosity with a dynamic explanation pattern was the LOE of choice, but
It is not in line with our results in  \textit{`with time limit'} condition.
Hence, future designers should keep in mind that the robot's explanation should be adaptive to the demands of the situation.
For both fluency and trust measures (see Fig.~\ref{fig:sub_without_time}), it was found that high/medium LOE was preferred over low LOE.
This result demonstrates that dynamic patterns are important with or without detailed explanations.

\subsection{Interview insights} \label{InterviewSection}
Several informative insights were derived from the interviews conducted with the participants after the experiment. 
First, we found that the interviews aligned with the results that we obtained by examining the effect of the LOE on the dependent variables. 
We found that in the \textit{`with time limit'} condition the level of verbosity, low vs. high, was not a significant factor but a dynamic pattern of explanation was required.
In contrast, in the \textit{`without time limit'} condition, a high level of verbosity and a dynamic pattern of explanation were preferred.

An interesting trend was found for the time-limited condition, namely, that the participants fell into two main groups, those preferring very detailed explanations and those preferring more general explanations. 
Members of the former group claimed that detailed explanations actually calmed them in stressful situations and hence reduced the burden and pressure on them as teleoperators, 
whereas those in the latter group argued that independent performance with the help of a more general or partially detailed explanation was preferable,
as it enabled superior task performance, without the hindrance of very detailed explanations (they claimed that a lot of information overloaded them and slowed down their performance). Therefore, there was no particular preference for high vs. medium LOE in the \textit{`without time limit'} condition, whereas the high LOE was clearly preferred in the \textit{`with time limit'} condition. In addition, in the \textit{`with time limit'} condition, there was also a higher percentage of participants who preferred the low LOE.

\subsection{Limitations}

The interaction and the evaluation of the study were conducted as a snapshot experiment in the lab. 
In general, the preferred LOE depends on many factors, such as the task (e.g., context/objective), the user (e.g., type, experience) and the environment. 
Each user might select a different LOE, and in addition, his/her preference for certain LOE might change along the task execution. 
In this work, the participants were unaware of the context of the task. This probably led them to prefer the  high LOE. 
As participants become more and more familiar with the task, they will not require a detailed explanation with a dynamic pattern. 
As demonstrated in~\cite{hellstrom2018understandable}, the explanations should be aligned to the discrepancy between the humans' perception of the robot's theory of mind and the robot's state of mind. 
For example, if the participants in our task (current study) were aware of the surroundings, then a shorter explanation with static patterns would have helped in navigating the robot.  
However, the alignment of discrepancy to the different levels of explanation needs to be explored in future studies. 

Additionally, this work focused only on a single task as a case study, namely, navigation in a simulated healthcare system.
Since it is important to implement HRI design on different types of tasks and robots~\cite{kumar2022politeness,ullman2021challenges}
the current design should be implemented on various collaborative tasks.
The specific LOE should be selected according to the task, the user, and the environment. 
For example, the robot would generate medium LOE in search-and-rescue operations that involve both task and time criticality. 
However, a manipulator robot collaborating with humans for an assembly task~\cite{kumar2024exploratory} needs to generate high LOE to provide a detailed explanation for each part of the assembly, to ensure accuracy and since the time is not critical.

We recruited students as a convenient sample, similar to many other studies in HRI that have explored their designs with students~\cite{chen2024effects,wang2024effects,washburn2020robot,camblor2024attention,kopp2023s} and to  enable to reach a larger sample, which is necessary to ensure statistical conclusions~\cite{hoffman2020primer}.
Furthermore, it ensured a homogeneous group, which is important so as not to introduce other factors that might influence the results (e.g., culture, age, background). 
In addition, we particularly sought to recruit students who have previous experience with robots on the assumption that technical users would be able to navigate successfully, even at a low LOE, vs. non-experienced users who would most probably require a high LOE. 
Since the focus of this paper was to evaluate different LOEs, it was important to ensure successful bypassing of obstacles for the comparison.
During the experiment, all the participants were able to navigate the robot in all three LOEs.  
Another reason for using students was that it enabled to avoid the novelty effect~\cite{reimann2023social}.
Users without previous robotic exposure (e.g., older adults) experience curiosity and excitement when encountering new technologies.
Hence, we decided to recruit students to enable the participants to focus on the research questions without being distracted by the technology or other effects.
 Future work should aim at extending the LOE model to additional populations.

\section{Conclusion and future work}
We have proposed a way to construct levels of explanation for designing instructions robots provide to humans. 
In this research, the explanation levels were based on two parameters, focusing on \textbf{What}, verbosity (high vs. low) and \textbf{When}, the explanation pattern (dynamic vs. static).
On the basis of these parameters, we proposed three LOE (high, medium, and low), which we compared in two time sensitive conditions, namely, \textit{`with time limit’} and \textit{`without time limit.’}

The LOE influenced all dependent variables. The high LOE was preferred (vs. the medium and low LOE) in terms of adequacy of explanation in the \textit{`without time limit’ condition}. 
Further, participants in high LOE made fewer collisions with obstacles, fewer wrong movements, and asked fewer questions to the experimenter.
However, completion time, fluency of interaction, and trust were similar for the medium and high LOE. 
In the \textit{`with time limit’} condition, it was found that there was no preference for the high over the medium LOE, but both were preferred to the low LOE. 
The explanation pattern had an effect on all
the dependent variables.
The dynamic pattern of explanation was helpful in both task conditions. 
The verbosity of explanation affected some of the variables (adequacy of explanation, number of collisions, wrong movements and number of clarifications) only in the \textit{`without time limit’} condition.
Hence, our study concurs with previous research~\cite{7451740} that found that a brief and precise explanation would be adequate for an emergency situation that is characterized by time limitation.

Future work should extend to other aspects of understanding (e.g., \textbf{Why}, \textbf{How}).
The methods proposed in this paper for LOE implementation and evaluation can serve as a basis for implementing additional aspects. 
For example, one important aspect to include is 
the justification of a robotic action (addressing the \textbf{Why}). 
If combined with  (\textbf{What}, and \textbf{When}) 
this would extend the LOE to 8 levels (corresponding to  3  questions).
However, for practical purposes, this is too complicated both for implementation and evaluation.
Hence, it might be worthy to design a hierarchical classification theme in which the LOE are dependent on context (for example LOE for \textbf{Why} will be included only in cases of errors or failures). 
To advance the field of understandable robots, future research should address different schemes along with their implementation and evaluation in user studies.
This could be extended in other types of tasks and environments (e.g., critical tasks such as emergency situations vs. service robots in both home and outdoor environments).
Future studies should also experiment with a diverse demographic of participants.

\section*{Declarations}
\subsection*{Acknowledgements}
This research was supported by Ben-Gurion University of the Negev through the Agricultural, Biological and Cognitive Robotics Initiative, the Marcus Endowment Fund, and the W. Gunther Plaut Chair in Manufacturing Engineering.
\subsection*{Data Availability}
The data set generated during and/or analyzed during the current study is available on \href{https://github.com/shikharkumar1993/Levels-of-explanations/tree/main}{Github}.
\subsection*{Conflict of interest}
The authors declare no conflict of interest.
\subsection*{Ethical Approval}
This study was approved by the ethical committee of the Department of Industrial Engineering and Management at Ben-Gurion University of the Negev.
\subsection*{Informed Consent}
Informed consent was obtained from all the subjects involved in the study.
\bibliography{lit}

\end{document}